\documentclass[10pt,twocolumn,letterpaper]{article}
 \usepackage{cvpr}
\usepackage[dvipsnames]{xcolor}

\definecolor{cvprblue}{rgb}{0.21,0.49,0.74}
\usepackage[pagebackref,breaklinks,colorlinks,citecolor=cvprblue]{hyperref}
\usepackage{times}  
\usepackage{helvet} 
\usepackage{courier}  
\usepackage[hyphens]{url}  
\usepackage{graphicx} 
\usepackage{natbib} 
\usepackage{caption}
\usepackage{multirow}
\usepackage{pifont}
\usepackage{amssymb}
\newcommand{\cmark}{\ding{51}}
\newcommand{\xmark}{\ding{55}}
\usepackage{algorithm}
\usepackage{algorithmic}
\usepackage{amsfonts}
\usepackage{amsmath}
\def\confName{CVPR}
\def\confYear{2024}

\title{MA-AVT: \underline{M}odality \underline{A}lignment for Parameter-Efficient \underline{A}udio-\underline{V}isual \underline{T}ransformers}
\author{
    Tanvir Mahmud\textsuperscript{\rm 1}, 
    Shentong Mo\textsuperscript{\rm 2},
    Yapeng Tian\textsuperscript{\rm 3},
    Diana Marculescu\textsuperscript{\rm 1} \\
    $^1$University of Texas at Austin, $^2$Carnegie Mellon University, $^3$University of Texas at Dallas
}
\begin{document}

\maketitle

\begin{abstract}
Recent advances in pre-trained vision transformers have shown promise in parameter-efficient audio-visual learning without audio pre-training. However, few studies have investigated effective methods for aligning multimodal features in parameter-efficient audio-visual transformers. In this paper, we propose MA-AVT, a new parameter-efficient audio-visual transformer employing deep modality alignment for corresponding multimodal semantic features. Specifically, we introduce joint unimodal and multimodal token learning for aligning the two modalities with a frozen modality-shared transformer. This allows the model to learn separate representations for each modality, while also attending to the cross-modal relationships between them. In addition, unlike prior work that only aligns coarse features from the output of unimodal encoders, we introduce blockwise contrastive learning to align coarse-to-fine-grain hierarchical features throughout the encoding phase. Furthermore, to suppress the background features in each modality from foreground matched audio-visual features, we introduce a robust discriminative foreground mining scheme. Through extensive experiments on benchmark AVE, VGGSound, and CREMA-D datasets, we achieve considerable performance improvements over SOTA methods. Code is released at \url{https://github.com/enyac-group/MA-AVT}.
\end{abstract}

\begin{figure}[t]
\centering
\includegraphics[width=0.98\columnwidth]{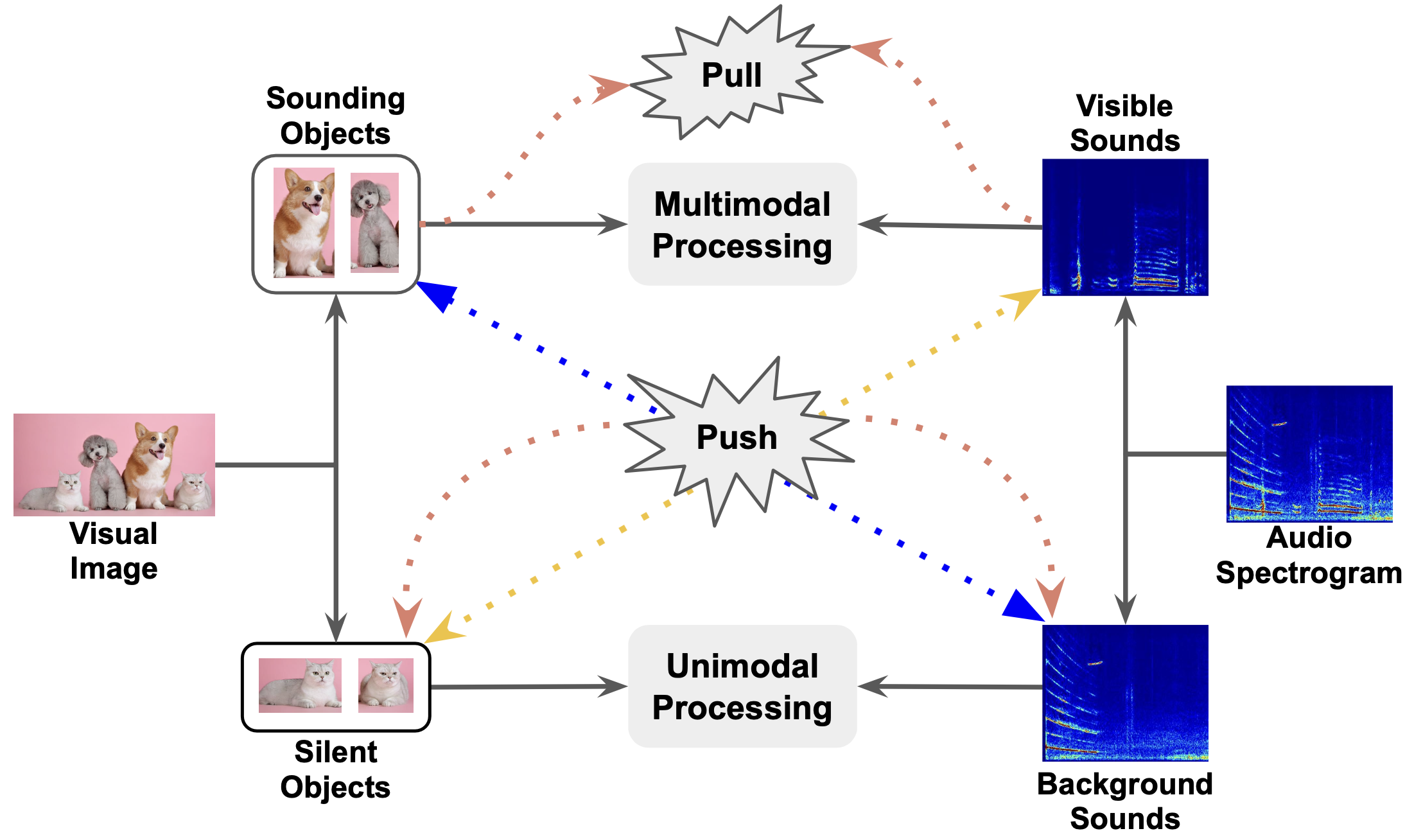} 
\caption{
A visual image contains sounding foreground object regions, as well as silent background regions. MA-AVT aims to align the foreground visual features with corresponding audio features. Simultaneously, MA-AVT learns mismatched uni-modal features to enhance cross-modal contrast.  In particular, MA-AVT leverages a pre-trained frozen vision transformer in audio-visual tasks with learnable uni-modal and shared cross-modal tokens.}
\label{f1}
\vspace{-4mm}
\end{figure}

\section{Introduction}
When observing a scene, humans can simultaneously identify sounding objects, distinguish background noises, and locate silent regions. Such natural audio-visual perception arises from fine-grain correspondence of environmental visual and auditory cues. In this work, we explore improving audio-visual feature alignment of audio-visual learners, from the initial feature extraction to the final decision-making stage, and systematically evaluate its impact on complex audio-visual recognition tasks.

Earlier work on audio-visual learning primarily focused on late multi-modal feature alignment by extracting features from separate uni-modal encoders \cite{psp, cmran, aveclip, avt}. These methods required large-scale separate audio and visual pre-training, as they used different encoders for each modality.
Later, to leverage deeper cross-modal fusion, \citet{mbt} introduced uniform audio and visual transformers with bottleneck fusion modules. This uniform architecture enabled blockwise fusion of cross-modal features across separate uni-modal encoders. 
However, this method requires large-scale joint training of uni-modal encoders with bottleneck fusion modules, which can be burdensome for deploying large transformer models (ViT-Large, 656M parameters).
Very recently, \citet{lavish} introduced a lightweight adapter module (LAVISH) to leverage pre-trained frozen vision transformer (ViT) in audio-visual tasks, without the need for any audio pre-training. 
This adapter enables deep cross-modal alignment by fusing the output of each ViT block features. Despite its promising performance and high parameter efficiency, this method has several limitations.

First, this method only trains cross-modal fusion modules with a frozen encoder, while ignoring unique unimodal feature components. However, natural images and sounds contain unique components that do not have multimodal correspondence but are still important for learning and separating the context. For example, most regions in visual frames usually contain silent objects, while the corresponding audio may contain significant background environmental sounds that are not present in the image. Therefore, it is necessary to simultaneously process unique unimodal and common cross-modal features in audio-visual learning.
To address this issue, we introduce joint unimodal and multimodal token learning, which simultaneously learns disjoint unimodal and common cross-modal representations. However, the learned tokens may have varying significance based on different class representations. To selectively enable the most relevant tokens in each modality, we introduce a local self-attention module for each group of tokens.

Second, LAVISH and most prior work~\cite{aveclip, guzhov2022audioclip} train only with late supervision applied to the coarse-grain features, extracted from the output of audio and visual encoders. This enforces explicit coarse-grain multimodal feature alignment, but the fine-grain features in earlier transformer blocks do not receive any explicit supervision for cross-modal alignment. To tackle this problem and align the hierarchical features of each modality throughout the encoding, we introduce blockwise semantic contrastive learning (SCL). Our approach inherently searches for the semantic visual representation with its corresponding audio representation by utilizing shared multimodal tokens. Thus, we introduce SCL on intermediate feature representations of multimodal tokens to strengthen modality alignment.

Third, audio and visual modalities may not contain any matching pairs of objects, resulting in complete misalignment. Employing fusion modules in such misaligned background regions can introduce significant noise during training.
However, existing audio-visual transformers, such as LAVISH~\cite{lavish} and MBT~\cite{mbt}, do not consider any background suppression methods to isolate such complete misalignment cases.
To overcome this limitation, we introduce a robust discriminative foreground mining scheme by learning additional background tokens that selectively suppress the mismatched background regions, thereby enhancing modality alignment.

\begin{figure*}[t]
\centering
\includegraphics[width=0.85\textwidth]{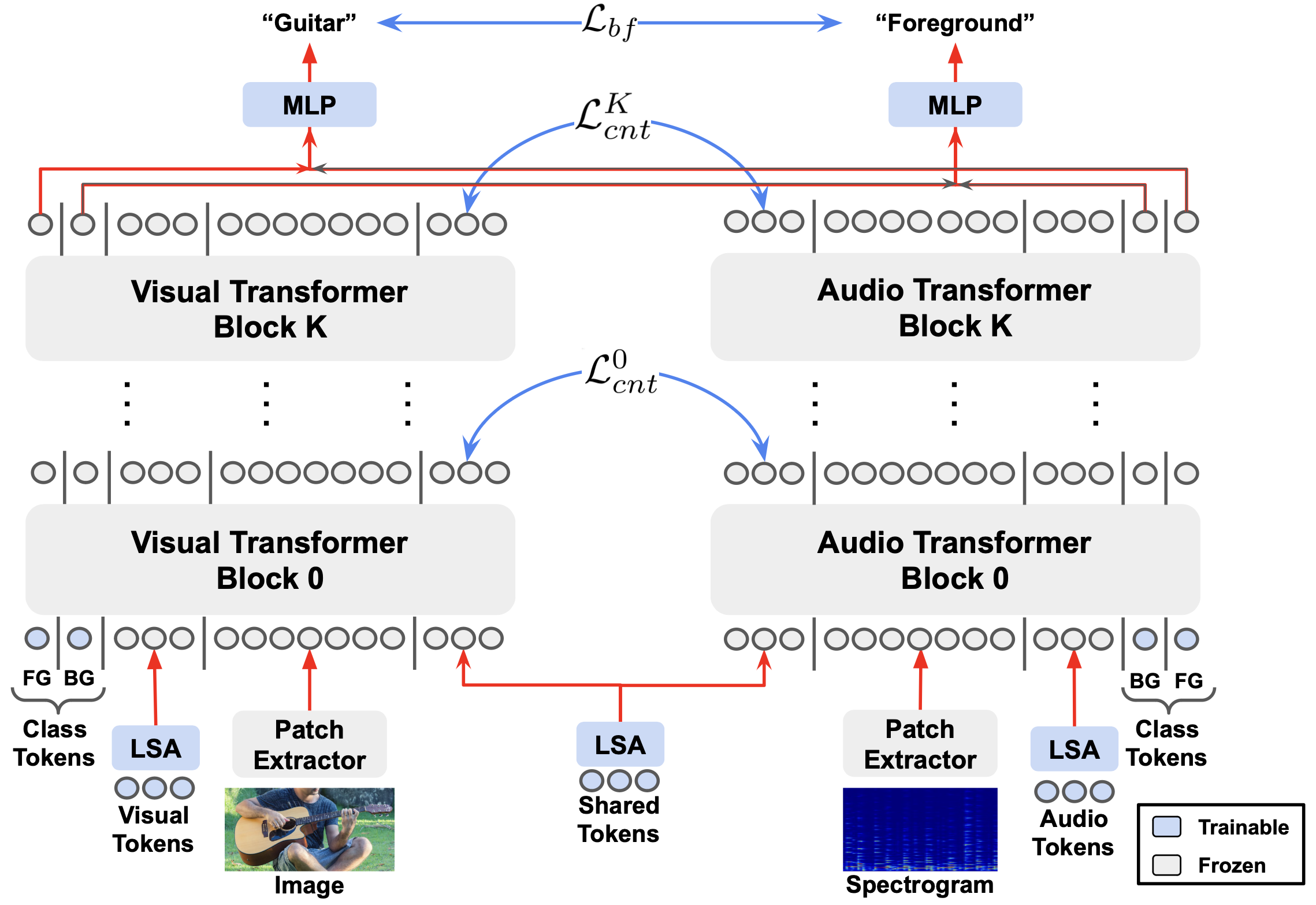}
\caption{\textbf{The overview of the proposed MA-AVT framework.} The image and audio spectrogram are processed simultaneously with frozen transformer encoders. Initially, we extract patch tokens using pre-trained patch extractors of transformers. We introduce learnable unimodal audio and visual tokens to learn unique unimodal representation as well as introduce multimodal shared tokens to learn joint representation. To focus on most relevant tokens for the target class, we introduce local self-attention (LSA) modules on each group of tokens. To further enhance the modality alignment, we operate blockwise semantic contrastive learning on the intermediate shared multimodal token embeddings after each transformer block. To suppress mismatching background regions, we introduce learnable background (BG) and foreground (FG) class tokens. Here, $\mathcal{L}_{bf}$ denotes foreground-background loss and $\mathcal{L}_{cnt}^k$ denotes contrastive loss after each $k^{th}$ block.
}
\label{fig2}
\end{figure*}

Extensive experiments on three popular benchmarks, \textit{i.e.} AVE, VGGSound, and CREMA-D datasets, demonstrate the superior performance of MA-AVT. 
The contributions of the proposed method can be summarized as follows:
\begin{enumerate}
    \item We propose joint unimodal and multimodal token learning for extracting disjoint uni-modal and common cross-modal features from each modality.

    \item We introduce blockwise semantic contrastive learning for deeper coarse-to-fine-grain hierarchical audio-visual feature alignment.

    \item We propose robust discriminative foreground mining to suppress the mismatched background features.

    \item Our method achieves significant performance improvements over other state-of-the-art approaches.
    
\end{enumerate}

\section{Related Works}
\subsection{Audio-Visual Learning}
In recent years, numerous approaches have been explored in diverse audio-visual learning applications, such as audio-visual source localization~\cite{mix, cnt1, cnt2, cnt3, easy}, sound separation~\cite{sop,owens2018audio,gao2018learning,xu2019recursive,gan2020music}, video parsing~\cite{parse1, parse2, parse3}, event localization~\cite{avel,aveclip, psp, mm}, and  classification~\cite{mbt, lavish, gradmod}. In this work, we primarily focus on designing parameter-efficient multi-modal transformer for the audio-visual classification task.

In general, much of earlier work in audio-visual learning has focused on the late fusion of extracted audio-visual features from separate pre-trained encoders. In MBT~\cite{mbt}, deep and mid-fusion approaches to audio-visual feature fusion have been explored with a unified transformer architecture with full tuning. However, these methods require large-scale pre-training on audio and visual data, which can be computationally expensive and time-consuming. Additionally, full-tuning can be parameter inefficient and prone to overfitting on smaller datasets, while partial tuning can result in sub-optimal results~\cite{vpt, lst, side}.
Recently, LAVISH~\cite{lavish} introduced pre-trained frozen vision transformers for audio-visual tasks without requiring any pre-training on audio data. However, LAVISH and other existing methods are trained with late supervision on the final projected feature spaces for audio-visual feature alignment. This can limit the effectiveness of feature alignment, as the final feature space may not capture all of the relevant information from the intermediate feature spaces.
In contrast, we study the impact of deep audio-visual feature alignment by leveraging supervision to the intermediate feature space of pre-trained frozen transformers with learnable tokens. This allows us to learn more robust and discriminative representations that are better aligned across modalities.

\subsection{Audio-visual Contrastive Learning}
Prior work explored audio-visual contrastive learning in self-supervised representation learning and sound source localization.  
\citet{global1} and \citet{global2} contrast across the global mean-pooled audio, and visual features.
However, in practice, audible objects correspond to a small portion of the image while audio includes background noise from non-visible objects. Hence, such global alignment introduces significant noise in contrastive learning. \citet{cnt1, cnt2, cnt3, easy, mil} contrast mean-pooled audio features with the most-similar corresponding image patch region. 
Nevertheless, the sounding object may extend to multiple patches across the image where these methods cannot represent the complete semantic relationships.
\citet{clip} used class token pooled features for representing visual modality. 
\citet{hard1} introduced hard positive sample mining in contrastive learning for semantic audio-visual matching. 
\citet{multi1} and \citet{avgn} introduced a combination of source classification and contrastive learning for multiple sounding source localization.
Most of the prior work mainly considered audio-visual correspondence on coarse-grain features extracted from the outputs of unimodal encoders.
Different from past work, we introduce audio-visual contrastive learning for aligning coarse-to-fine-grain hierarchical features in audio-visual transformers.

\section{Methodology}
Our goal is to build a parameter-efficient audio-visual transformer with modality alignment for audio-visual recognition tasks. The overview of the proposed framework is shown in Figure~\ref{fig2}. To maintain learning efficiency, we adopt frozen ViTs as audio and visual encoders.
Upon the framework, we first extract corresponding patch token embeddings from the input image and audio spectrogram. Then, we introduce unimodal token embeddings to both modalities to learn separate and unique unimodal feature representations. We also introduce shared multimodal tokens to both modalities to learn the common cross-modal semantic relationship.
To focus on the most relevant tokens for the target class, we utilize local self-attention (LSA) modules for each group of tokens. In addition, we propose to use background tokens to detect complete mismatches of audio-visual pairs and foreground class tokens to detect the audio-visual class in both modalities.
After aggregating all token embeddings separately for each modality, we process frozen transformer blocks sequentially. The shared multimodal tokens inherently search for the audio-visual corresponding regions in both modalities.
To further strengthen cross-modal alignment, we introduce blockwise semantic contrastive learning (SCL) across the mean-pooled intermediate token representations extracted from each transformer block. This blockwise contrastive learning is only used during training to align coarse-to-fine-grain audio-visual feature representations.
Finally, the foreground class token generates the foreground class prediction, and the background class token predicts the binary background class. In the case of complete misaligned background class predictions, the foreground and blockwise contrastive loss gradient propagations are suppressed during training.

\subsection{Preliminaries}
Let's consider the dataset $\mathcal{D} = \{ (v_i, a_i): i = 1, \dots, N\}$ representing $N$ pairs of images $v_i \in \mathbb{R}^{3 \times H \times W}$ sampled from a video at time $t$, and corresponding audio spectrograms $a_i \in \mathbb{R}^{F \times T}$ centered at time $t$ with a span of several seconds. 
Following previous work~\cite{lavish, mbt}, $v_i$ is divided into $m$ non-overlapping patches which are flattened to extract image patch token embeddings $P_v^{0} \in \mathbb{R}^{m \times d}$. Similarly, audio spectrogram $a_i$ is split into $n$ audio patch token embeddings $P_a^{0} \in \mathbb{R}^{n \times d}$. Moreover, linear interpolation across pre-trained position embeddings is carried out, particularly for audio tokens, to match with the number of patch tokens in case of mismatch with pre-trained embeddings.

\subsection{Multimodal Alignment with Learnable Tokens}
Leveraging a pre-trained frozen vision transformer in audio-visual learning~\cite{lavish} poses two main challenges.
First, the disparate unimodal audio and visual features should be extracted separately since visual and auditory signals come with unique feature representations. Second, the grounded modality-shared features across two modalities need to be processed simultaneously as a fusion step. To solve the former one, we introduce \textbf{modality-specific} token prompts which are uniquely trained for each modality. And, for the latter one, we introduce \textbf{modality-invariant} token prompts which are jointly trained across both modalities. Before merging with the patch token embeddings, all three bags of modality-invariant and modality-specific tokens are processed with separate local self-attention (LSA) units. These units enhance vanilla prompt training capacity by focusing on most relevant prompts for different classes. These LSA units are formed with residual multi-headed attention (MHA)~\cite{transformer} operations such that,
\begin{equation}
    LSA(x) = x + \text{MHA}(x)
\end{equation}
Let's denote LSA units operating on audio, visual, and multimodal shared prompt tokens as $A_a (\cdot)$, $A_v (\cdot)$, and $A_s (\cdot)$, respectively, and bags of audio, image, and shared tokens as $z_a \in \mathbb{R}^{n_a \times d}$,
$z_v \in \mathbb{R}^{n_v \times d}$, and $z_s \in \mathbb{R}^{n_s \times d}$, respectively. Hence, intermediate image and audio token embeddings $(E_a, E_v)$ after $k^{th}$ transformer block are given by
\begin{align}
    E_a^k = A_a(z_a)^k\ ||\ P_a^k \ || \ A_s({z}_s)^k; \ \ \ \forall k = \{1, \dots, K \} \\
    E_v^k = A_v(z_v)^k\ ||\ P_v^k\ ||\ A_s({z}_s)^k; \ \ \ \forall k = \{1, \dots, K \}
\end{align}
where $K$ denotes a total number of transformer blocks, and $||$ denotes the feature concatenation operation.

\subsection{Blockwise Semantic Contrastive Learning}
In practice, visual frames consist of target foreground sounding regions as well as silent background regions. Similarly, audio contains target foreground sounding sources along with certain noises from invisible background sounding sources. Proper understanding of audio-visual scenes poses two primary challenges. 
First, the model should properly align the corresponding semantic regions of audio-visual features representing high cross-modal similarity as well as distinguish unique unimodal features. Second, the model should discriminate among hierarchical audio-visual features of target classes.

Our unimodal and multimodal shared prompting technique inherently solves the first problem. Nevertheless, we introduce semantic contrastive learning to further strengthen the modality alignment. Notably, the shared tokens explicitly search for the semantic regions with high audio-visual correspondence over all other patch tokens in each modality. Hence, we extract the semantic representation of the target matching pair by taking the mean over the output of shared token embeddings in each modality. We introduce cross-modal contrastive learning over these mean pooled semantic feature representations, which is denoted as semantic contrastive learning (SCL).

The supervised learning objective operating on the coarse-grain audio-visual features inherently generates discriminative hierarchical features in subsequent layers, thereby attempting to solve the second problem. Along with such coarse-grain supervision, we introduce blockwise semantic contrastive learning for further alignment of the fine-grain hierarchical features. We note that such blockwise cross-modal alignment doesn't alter the inter-block hierarchical feature relationships generated with supervised learning. Rather, it generates deeper auxiliary supervision throughout the encoding phase to contrast across coarse-to-fine-grain cross-modal semantic relationships.

After $k^{th}$ block, assume $a^k = \frac{1}{n_s}\sum z_{s,a}^k \in \mathbb{R}^{1\times d}$ and $v^k = \frac{1}{n_s}\sum z_{s,v}^k\in \mathbb{R}^{1\times d}$ represent the mean-pooled shared token features from audio, and visual modality, respectively. The blockwise semantic contrastive loss ($\mathcal{L}_{cnt}^k$) at $k^{th}$ block with a batch size of $B$ is given by
    
\begin{align}
&    \mathcal{L}_{cnt}^k = (\mathcal{L}_{v \rightarrow a}^k + \mathcal{L}_{a \rightarrow v}^k)/2 \\
&    \mathcal{L}_{v \rightarrow a}^k = -\frac{1}{B} \sum\limits_{b=1}^B \log \frac{ \exp(\frac{1}{\tau}sim( v^k_b, a^k_b))}{\sum\limits_{j=1}^B exp(\frac{1}{\tau}sim(v^k_b, a^k_j))} \\
&    \mathcal{L}_{a \rightarrow v}^k = -\frac{1}{B} \sum\limits_{b=1}^B \log \frac{ \exp(\frac{1}{\tau}sim( a^k_b, v^k_b))}{\sum\limits_{j=1}^B exp(\frac{1}{\tau}sim(a^k_b, v^k_j))} 
\end{align}
where $sim(v^k, a^k)$ represents the cosine similarity, and $\tau$ is a temperature parameter.

\subsection{Robust Discriminative Foreground Mining}
Audio may contain only background noise or sound from non-visible objects, which results in a complete misalignment of audio-visual features. We denote such cases as \textit{background class}. If the audio-visual pair contains at least one matching object, that will represent the \textit{foreground class}. In certain datasets (\textit{e.g.}, VGGSound~\cite{vggsound}) where we don't have access to mismatched pairs, we consider synthetic audio-visual pairs from two separate sources for learning such mismatched background cases.

We introduce unique background class tokens $z_b \in \mathbb{R}^{1 \times d}$ and foreground class tokens $z_f \in \mathbb{R}^{1 \times d}$ in both modalities. Thus, the accumulated tokens at $k^{th}$ layer are:
\begin{align}
    E_a^k = z_b^k \ ||\ A_a(z_a)^k\ ||\ P_a^k \ || \ A_s({z}_s)^k\ ||\ z_f^k \\
    E_v^k = z_b^k \ ||\ A_v(z_v)^k\ ||\ P_v^k\ ||\ A_s({z}_s)^k\ ||\ z_f^k
\end{align}
The output of these class tokens from the last transformer blocks are concatenated, and processed with MLP layers to generate final background prediction $y_b \in \mathbb{R}^{1\times 1}$ and foreground prediction $y_f \in \mathbb{R}^{1\times C}$. The foreground-background loss ($\mathcal{L}_{bf}$), and total loss ($\mathcal{L}_{total}$) are given by
\begin{align}
&    \mathcal{L}_{bf} = y_b * BCE(\hat{y}_b, y_b) + (1 - y_b)  * CE(\hat{y}_f, y_f) \\
&    \mathcal{L}_{total} =  \mathcal{L}_{bf} + (1 - y_b)  * \sum_{k=1}^K \mathcal{L}_{cnt}^k
\end{align}
where $y_b = \{0, 1 \}$ and $y_f = \{0, 1, \dots, C \}$ represent the binary background label and C-class foreground labels, respectively. The foreground  and semantic contrastive losses are suppressed for mismatched pairs.

\section{Experiments}

\subsection{Datasets}
\paragraph{AVE~\cite{avel}} dataset contains $4,143$ videos of $10$-second audio and visual segments. It has per-second frame-level annotations for audio-visual event localization and consists of $28$ event classes along with background class annotations representing complete modality misalignment. This dataset has natural misaligned audio-visual pairs that makes it directly applicable to MA-AVT. The data-split contains $3,942$ training videos, $742$ test videos, and $892$ validation videos. Following prior work, we sample image frames at $1$ fps, and extract corresponding audio segments of $1$s duration. We also use the same evaluation metric of the fraction of correctly predicted event regions as in prior work~\cite{avel}.
\vspace{-2mm}

\paragraph{VGGSound~\cite{vggsound}} is a large-scale audio-visual learning dataset containing $309$ classes that represent ``in the wild" audio-visual correspondence. This dataset contains a large range of sounding events from day-to-day life. All videos are collected from YouTube. Since many videos are not available anymore, we use $161,234$ videos for training, and $12,873$ videos for testing following the data split in~\cite{vggsound}. For training and evaluation, we sample single image frames per video from the middle of each video and extract corresponding audio segments of $5$s duration. We use the same evaluation metric of class accuracy following prior work. Since the dataset size is reduced for unavailable videos, we reproduced the reported results of prior work under the same settings for fair comparison.
Since this dataset only contains audio-visual matched pairs, we introduce synthetic mis-matched pairs during training, particularly to learn foreground-background tokens.

\vspace{-2mm}

\paragraph{CREMA-D~\cite{crema}} is a speech emotion recognition dataset with $7,442$ video clips of $2 \sim 3$ seconds duration collected from $91$ actors. Each actor speaks various short words with $6$ usual emotion categories, such as \textit{happy, sad, angry, neutral, disgust, discarding, and fear}. The dataset is annotated by crowd-sourcing from $2,443$ raters for categorical emotion labels. We use the same train and test split as prior work~\cite{crema}, more specially $6,698$ training videos, and $744$ test videos. 
We use a single frame sampled from the middle of the video and corresponding audio segments with $3$s duration. Similar to prior work, we use the same evaluation metric of emotion recognition accuracy. We introduce similar synthetic mismatched audio-visual pairs during training.

\subsection{Implementation Details}
We use the spectrogram representation for audio by repeating the channel from $1$ to $3$ to operate with the same ViT backbone. All image samples are resized to $(224, 224)$. All audio spectrograms are extracted with a window length of $512$ and overlap of $353$. We use $5$ tokens for all audio, visual, and shared multimodal cases in all experiments unless otherwise specified. We use ADAM optimizer with the initial learning rate of $1e-3$ which is multiplied by $0.1$ after every $30$ epochs. All models are trained on $4$ A5000 GPUs with 24GB memory. We use a batch size of $256$ for training all models.

\begin{table*}[]
\caption{\textbf{Comparison with state-of-the-art methods.} We present comparison with CNN and transformer-based approaches. For the audio encoder, we either use audio pre-trained weights from Audioset or simple image-pretrained weights from ImageNet dataset. * denotes our improved implementation. Other than the reported results on AVE, we reproduced the results on VGGSound and CREMA-D datasets from open source implementation. The parameter counts are calculated for AVE dataset. (T) represents fully-trainable and (F) represents frozen encoder. Our proposed MA-AVT achieves significant performance improvements compared to other state-of-the-art methods while maintaining parameter-efficiency.}
\label{t1}
\centering
\scalebox{0.81}{
\begin{tabular}{lccccccccc}
\toprule
\multirow{2}{*}{Method} & \multicolumn{2}{c}{Visual Encoder} & \multicolumn{2}{c}{Audio Encoder} & \multirow{2}{*}{\begin{tabular}[c]{@{}c@{}}Total \\ Params (M)\end{tabular}} & \multirow{2}{*}{\begin{tabular}[c]{@{}c@{}}Trainable\\ Params (M)\end{tabular}} & \multicolumn{3}{c}{Accuracy (\%)}       \\
\cmidrule(lr){2-3} \cmidrule(lr){4-5} \cmidrule(lr){8-10}
                        & Model           & Pretrain   & Model         & Pretrain   &                                                                             &                                                                                 & VGGSound\footnotemark  & AVE           & CREMA-D \\
                        \midrule
PSP~(\cite{psp})                     & VGG-19 (F)        & ImageNet        & VGGish (F)      & AudioSet        & 231.5                                                                       & 1.7                                                                             & 50.1        & 77.8          & 62.5       \\
AVEL~(\cite{avel})                    & ResNet-152 (F)    & ImageNet        & VGGish (F)      & AudioSet        & 136.0                                                                       & 3.7                                                                             & 48.3        & 74.0          & 59.7       \\
CMRAN(~\cite{cmran})                   & ResNet-152 (F)    & ImageNet        & VGGish (F)      & AudioSet        & 148.2                                                                       & 15.9                                                                            & 50.7        & 78.3          & 61.8       \\
MM-Pyramid~(\cite{mm})              & ResNet-152 (F)    & ImageNet        & VGGish (F)      & AudioSet        & 176.3                                                                       & 44.0                                                                            & 49.8        & 77.8          & 63.1       \\ \midrule
MBT~(\cite{mbt})                    & ViT-B-16 (T)      & ImageNet        & ViT-B-16 (T)    & ImageNet        & 206.4                                                                       & 206.4                                                                           & 54.6        & 76.1             & 70.7       \\
LAVISH~(\cite{lavish})                 & ViT-B-16 (F)      & ImageNet        & ViT-B-16 (F)    & ImageNet        & 107.2                                                                       & 4.7                                                                             & 52.5        & 75.3          & 68.9       \\
LAVISH*~(\cite{lavish})                 & ViT-B-16 (F)      & ImageNet        & ViT-B-16 (F)    & ImageNet        & 110.4                                                                       & 7.3                                                                             & 53.6        & 75.8          & 69.7       \\
\textbf{MA-AVT (ours)}  & ViT-B-16 (F)      & ImageNet        & ViT-B-16 (F)    & ImageNet        & 110.2                                                                       & 7.1                                                                             & \textbf{56.7}        & \textbf{77.9} & \textbf{72.3}       \\ \midrule
MBT*~(\cite{mbt})                     & ViT-L-16 (T)      & ImageNet        & ViT-L-16 (T)    & ImageNet        & 656.8                                                                           & 656.8                                                                               & 57.1        & 78.8             & 72.4       \\
LAVISH~(\cite{lavish})                  & ViT-L-16 (F)      & ImageNet        & ViT-L-16 (F)    & ImageNet        & 340.1                                                                       & 14.5                                                                            & 55.4        & 78.1          & 70.3       \\
\textbf{MA-AVT (ours)}  & ViT-L-16 (F)      & ImageNet        & ViT-L-16 (F)    & ImageNet        & 338.4                                                                       & 12.6                                                                             & \textbf{58.6}        & \textbf{79.6} & \textbf{74.9}  \\ \midrule
MBT*~(\cite{mbt})                     & ViT-B-16 (T)      & ImageNet        & AST (T)         & AudioSet        & 172                                                                         & 172                                                                             & 56.1        & 77.8          & 73.8
      \\
\textbf{MA-AVT (ours)}  & ViT-B-16 (F)      & ImageNet        & AST (F)         & AudioSet        & 180.3                                                                       & 8.3                                                                             & \textbf{59.1}        & \textbf{80.3} & \textbf{75.2}  \\ 
\bottomrule   
\end{tabular}}
\vspace{-2mm}
\end{table*}

\subsection{Experimental Study}
In this work, we propose enhanced modality alignment techniques for parameter efficient audio-visual transformers. To demonstrate the effectiveness of the proposed method, we compare the performance with state-of-the-art approaches on three popular benchmark datasets. For fair comparison, we reproduced the results of most other approaches from their open-sourced implementation under the same setting.  We also present the ablation study to show the effectiveness of different building blocks. We use VGGSound dataset and ViT-B-16 backbone for most of the ablations unless otherwise specified.

\footnotetext{Since the dataset size in VGGSound is considerably reduced for unavailable videos, we report the reproduced result in the same setup.}

\vspace{-3mm}

\paragraph{Comparison with prior CNN-based approaches:}
As shown in Table~\ref{t1}, we study the performance of several recent CNN-based approaches~\cite{avel, psp, cmran, mm}. Most of these methods rely on separate pre-trained audio and visual encoders for the feature extraction with different late-fusion techniques. Despite the use of late-fusion techniques for modality alignment, the early stage of feature extraction only relies on unimodal encoders. Hence, the fusion techniques are only operated on the coarse unimodal feature representation thereby limiting performance. Moreover, performance of these methods depends on pre-training on large-scale audio and image data.  Our method achieves $+1.3$, $+7.9$, and $+11.8$ accuracy improvements in AVE, VGGSound, and CREMA-D datasets, respectively, compared to the best-performing CNN-based counterparts without using any audio pre-training. 

\paragraph{Comparison with prior transformer-based works:}
We also compare with several state-of-the-art transformer-based methods for audio-visual tasks~\cite{lavish, mbt}. Compared to CNN based methods, transformer based approaches can leverage early fusion techniques for using the uniform architecture in both modalities. As shown in Table~\ref{t1}, our proposed MA-AVT achieves superior performance compared to other existing transformer based methods. We primarily focus on the recently released LAVISH~\cite{lavish} and MBT~\cite{mbt} models for the discussion. 
 
LAVISH~\cite{lavish} introduces a parameter-efficient vision transformer based network for audio-visual tasks without using audio pre-training. We achieve $+2.6, +4.2, +3.4$ accuracy improvements on AVE, VGGSound, and CREMA-D datasets, respectively, compared to LAVISH for the ViT-B-16 model. We also demonstrate the performance improvements with other ViT architectures (ViT-L-16). 
We note that the number of additional trained parameters in MA-AVT is comparable with LAVISH (\textbf{7.1M} vs. \textbf{4.7M} in ViT-B-16). For fair comparison, we increase the number of trained parameters in LAVISH by using larger convolutions in adapter modules  ($\sim$ \textbf{7.3}M). Nevertheless, our method outperforms the larger LAVISH model with a considerable margin.

We hypothesize these performance improvements of MA-AVT are due to three significant modifications.  First, with a pre-trained frozen vision encoder as a common backbone for both modalities, LAVISH only trains cross-modal fusion adapters at each block. However, both audio and vision modalities have significant unique unimodal feature components which can be suppressed by focusing only on fusion adapters. In contrast, we adapt to significant unique unimodal and common cross-modal feature components by leveraging separate unimodal and multimodal tokens with local self-attention modules.
Second, despite using early fusion techniques, LAVISH only considers supervision on the coarse features extracted from unimodal encoder outputs. 
In contrast, our blockwise semantic contrastive learning in MA-AVT operates on hierarchical coarse-to-fine-grain features thereby resulting in deeper alignment of multimodal features. Third, we introduce robust foreground mining to suppress the complete misalignment of background audio-visual pairs, which is absent in LAVISH.

\begin{table}[t]
\centering
\caption{\textbf{Effect of unimodal and multimodal learnable Tokens with local self attention (LSA).} Combination of audio, visual, and shared multimodal tokens generates best performance. Integration of LSA module considerably improves the performance. Only classification loss is used for this analysis.}
\label{t3}
\scalebox{0.9}{
\begin{tabular}{ccccc}
\toprule
\multirow{2}{*}{\begin{tabular}[c]{@{}c@{}}Audio\\ Tokens\end{tabular}} & \multirow{2}{*}{\begin{tabular}[c]{@{}c@{}}Visual\\ Tokens\end{tabular}} & \multirow{2}{*}{\begin{tabular}[c]{@{}c@{}}Shared\\ Tokens\end{tabular}} & \multirow{2}{*}{\begin{tabular}[c]{@{}c@{}}Accuracy (\%)\\ w/o LSA\end{tabular}} & \multirow{2}{*}{\begin{tabular}[c]{@{}c@{}}Accuracy (\%)\\ with LSA\end{tabular}} \\
                              &                                &                                &                                                                                  &                                                                                   \\ \midrule
\cmark                             & \xmark                              & \xmark                              & 43.3                                                                             & 46.8                                                                              \\
\xmark                             & \cmark                              & \xmark                              & 41.5                                                                             & 45.7                                                                              \\
\xmark                             & \xmark                              & \cmark                              & 45.4                                                                             & 49.3                                                                              \\
\cmark                             & \cmark                              & \xmark                              & 48.5                                                                             & 52.5                                                                              \\
\cmark                             & \cmark                              & \cmark                              & \textbf{51.2}                                                                    & \textbf{54.1}                                                                    \\ \bottomrule
\end{tabular}}
\end{table}

\begin{table}[]
\centering
\caption{\textbf{Comparison with existing contrastive learning methods.} $V_{xy}$ denotes the visual patch embedding of position $(x, y)$ and $A$ denotes the mean audio patch embedding. $A_{cls}$, $V_{cls}$ denote the output of corresponding class token embedding. $V_{s}, A_{s}$ denote the output of shared multimodal token embeddings.}
\label{t2}
\scalebox{0.9}{
\begin{tabular}{ccc}
\toprule
Position                                                               & Matching Function &  VGGSound  \\ \midrule
\multirow{4}{*}{\begin{tabular}[c]{@{}c@{}}Final\\ Block\end{tabular}} & $\text{MaxPool}_{xy}(\text{sim}(V_{xy}, A))$~\cite{easy}     & 54.7         \\
                                                                       & $\text{AvgPool}_{xy}(\text{sim}(V_{xy}, A))$~\cite{mil}       & 54.3         \\
                                                                       & $\text{sim}(\text{MaxPool}_{xy}(V_{xy}), A)$~\cite{cnt2}       & 54.1          \\
                                                                       &  $\text{sim}(V_{cls}, A_{cls})$~\cite{clip} & {54.0}         \\
                                                                       &  $\text{sim}(\text{Mean}(V_{s}), \text{Mean}(A_{s}))$~(Ours)      & \textbf{55.6}          \\ \midrule
\multirow{4}{*}{\begin{tabular}[c]{@{}c@{}}Block-\\ wise\end{tabular}} & $\text{MaxPool}_{xy}(\text{sim}(V_{xy}, A))$~\cite{easy}     & 55.8         \\
                                                                       & $\text{AvgPool}_{xy}(\text{sim}(V_{xy}, A))$~\cite{mil}     & 55.2         \\
                                                                       & $\text{sim}(\text{MaxPool}_{xy}(V_{xy}), A)$~\cite{cnt2}     & 54.9         \\
                                                                       & $\text{sim}(V_{cls}, A_{cls})$~\cite{clip} & 55.1       \\
                                                                       & $\text{sim}(\text{Mean}(V_{s}), \text{Mean}(A_{s}))$~(Ours)     & \textbf{56.7}      \\ \bottomrule
\end{tabular}}
\vspace{-2mm}
\end{table}

In MBT~\cite{mbt}, modality bottleneck fusion tokens are introduced with separate audio and vision transformers with full fine-tuning of transformer encoders. We show the performance of MBT with both ImageNet pretrained weights and AudioSet pretrained weights particularly for the audio encoder. Our method achieves $+1.8, +2.1, +1.6$ higher accuracy than ImageNet-pretrained MBT with ViT-B-16 encoder on AVE, VGGSound, and CREMA-D datasets, respectively.
These improvements demonstrate that our proposed MA-AVT provides superior results to fully-tuning transformer encoders despite being trained on significantly smaller number of parameters (\textbf{7.1M} vs. \textbf{206.4M}). With AudioSet pre-trained weights in audio encoder, MA-AVT achieves $+2.5, +3.0, +1.4$ higher accuracy on AVE, VGGSound, and CREMA-D benchmarks, respectively. 
In general, audio encoders with AudioSet pretrained weights achieve higher performance than ImageNet pretrained weights for large-scale audio pretraining.
Nevertheless, our approach achieves consistent performance improvement over MBT. We hypothesize that these improvements are due to our blockwise semantic contrastive learning for deeper modality alignment and our robust foreground mining methods to suppress background pairs, both of which are missing in MBT.

\begin{table}[t]
\caption{\textbf{Effect of robust foreground (FG) mining.} Integration of robust FG mining considerably improves the performance by learning complete mismatch cases.}
\label{t4}
\centering
\scalebox{1.0}{
\begin{tabular}{lccc}
\toprule
Method   & AVE  & VGGSound & CREMA-D \\ \midrule
w/o FG Mining & 77.1 & 55.6     & 71.4    \\
with FG mining                  & \textbf{77.9} & \textbf{56.7}     & \textbf{72.3}   \\ \bottomrule
\end{tabular}}
\end{table}

\vspace{-2mm}

\begin{table}[t]
\centering
\caption{\textbf{Effect of multimodal alignment on unimodal learning.} Our multi-modal alignment method improves uni-modal accuracy.}
\label{t5}
\scalebox{1.0}{
\begin{tabular}{ccc}
\toprule
\multirow{2}{*}{Modality} & \multirow{2}{*}{\begin{tabular}[c]{@{}c@{}}multimodal\\  Alignment\end{tabular}} & \multirow{2}{*}{\begin{tabular}[c]{@{}c@{}}Accuracy\\ (\%)\end{tabular}} \\
                          &                                                                                   &                                                                          \\ \midrule
Audio                     & \xmark                                                                               & 40.2                                                                     \\
                          & \cmark                                                                              & \textbf{43.3}                                                                     \\ \midrule
\multirow{2}{*}{Visual}   & \xmark                                                                               & 38.7                                                                     \\
                          & \cmark                                                                              & \textbf{41.5}                                                                   \\ \bottomrule
\end{tabular}}
\vspace{-2mm}
\end{table}

\paragraph{Effect of unimodal and multimodal tokens with local self-attention (LSA) modules:}
In Table~\ref{t3}, we ablate the effect of unimodal and multimodal tokens with LSA modules in MA-AVT. Only classification loss is used for this analysis.
We note that audio-only tokens provide $+1.8$ higher accuracy than image-only tokens without using any LSA modules. Since we only use single image frame per video with full-length audio, the audio contains richer details of the sounding event compared to single images. Combining unimodal audio and visual tokens provides $+5.2$ higher accuracy than the audio-only tokens. This shows the combination of audio and visual tokens performs better than single modal tokens. By only using shared multimodal tokens, we achieve $+2.1$ higher accuracy than the unimodal audio only tokens. Shared tokens are supposed to perform multimodal alignment over audio and visual modality that can be responsible for the accuracy improvement. Finally, the combination of audio, visual, and multimodal tokens achieves the best accuracy of $51.2$ which is $+5.8$ higher than shared multimodal tokens and $+2.7$ higher than unimodal combinations. Finally, we note consistent performance improvements with the incorporation of local self-attention (LSA) modules. We achieve the highest accuracy of $54.1$ by integrating LSA modules in all three tokens which demonstrates the effectiveness of our approach.

\vspace{-2mm}
\paragraph{Comparison with existing contrastive learning methods:}
In Table~\ref{t2}, we present the results of various contrastive learning methods. For fair comparison, we keep the same architecture of MA-AVT for all contrastive losses. We separately present the effect of blockwise contrastive loss for all these cases. 
We note that our proposed semantic contrastive loss provides significant performance improvements for all benchmarks. Moreover, we also observe consistent performance improvements with the incorporation of the blockwise contrastive losses in all cases.
Prior work mostly focuses on matching the spatial visual features $V = \{V_{xy}: \forall x,y \}$ from all $(x, y)$ position with the global mean of audio features $A$ \cite{cnt2, easy, mil}. 
Considering the complex semantic relations with audio and visual modality in the presence of visual backgrounds and secondary noisy-sounding sources, most of these approaches introduce significant noise in training. 
In addition, contrastive learning applied on class-token provides performance inferior to ours~\cite{clip}. 
In contrast, our approach searches for the corresponding semantic regions in both audio and visual modalities simultaneously by utilizing the shared multimodal token embeddings which further strengthens the modality alignment.

\begin{figure}[t]
\centering
\includegraphics[width=0.95\columnwidth]{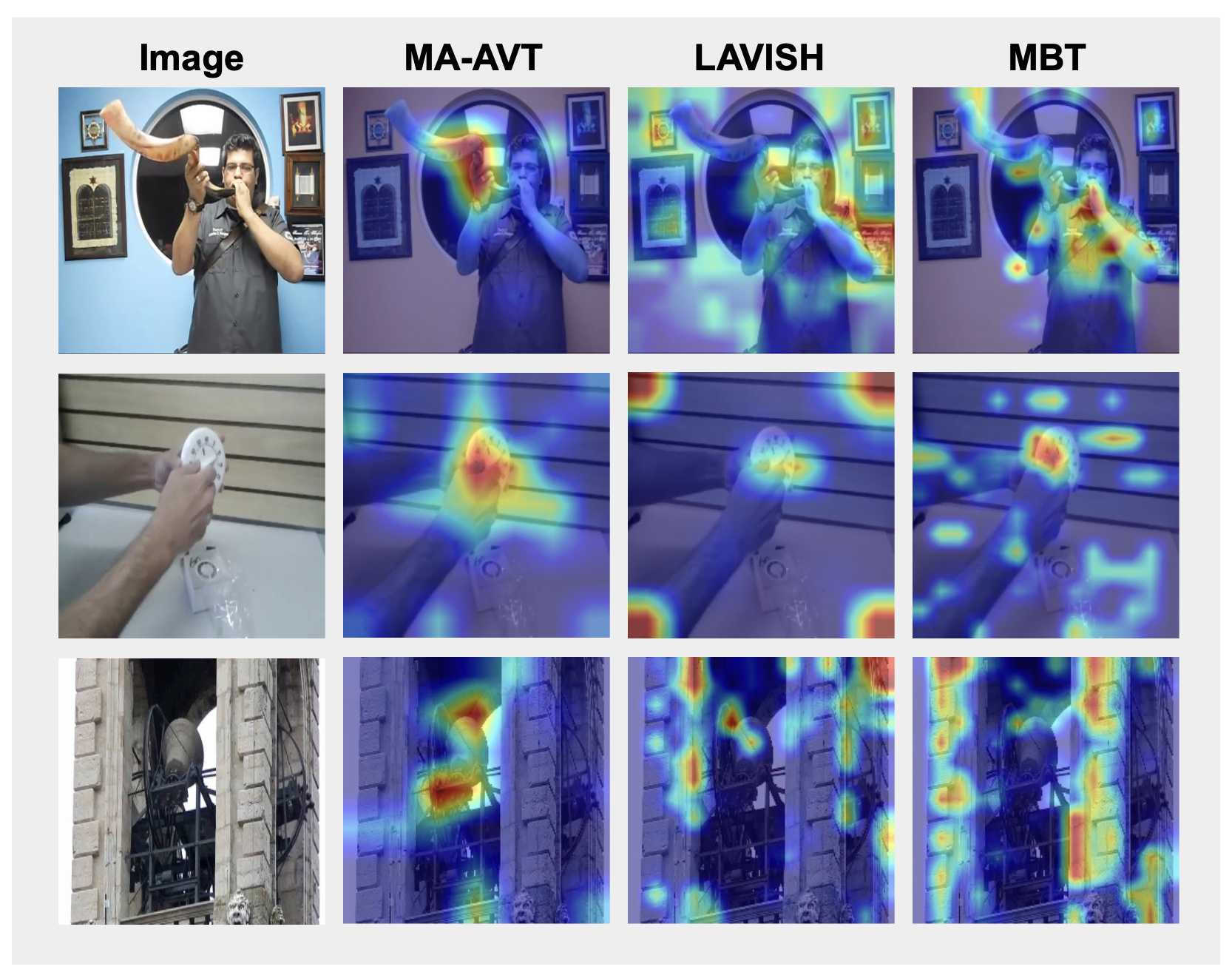}
\vspace{-3mm}
\caption{\textbf{Grad-CAM visualization for qualitative comparison.} Here, red color denotes high attention values and blue color denotes low attention values. Modality alignment brings noticeable improvements in MA-AVT to put more attention on the sounding regions. In general, MA-AVT better discovers the target visual sounding regions with sharper boundaries compared to other competitive baselines. Moreover, MA-AVT significantly reduces the attention weights on the silent regions.}
\label{sf1}
\vspace{-5mm}
\end{figure}

\paragraph{Effect of robust discriminative foreground mining:}
We ablate the effect of robust foreground mining as shown in Table~\ref{t4}. The AVE dataset contains annotations for background classes that can be directly used for both training and testing of the proposed method. However, for the VGGSound, and CREMA-D datasets, we use the robust foreground mining only in training, where we randomly choose audio and images from two different classes to represent the background class. By integrating foreground mining, we achieve $+0.8$, $+1.1$, and $+0.9$ higher accuracy on AVE, VGGSound, and CREMA-D datasets, respectively.

\paragraph{Multimodal alignment helps unimodal learning:}
MA-AVT can work on unimodal data in test scenarios by only using the separate unimodal tokens and by splitting the last MLP layer for foreground classification symmetrically. 
In Table~\ref{t5}, we study the effect of multimodal alignment methods on unimodal learning.
We achieve $+3.1$, $+2.8$ higher accuracy on audio and visual modalities, respectively, when comparing the proposed multimodal alignment techniques with separate unimodal training. We hypothesize the multimodal alignment helps unimodal learning by searching for the target region-of-interests in both modalities.

\section{Qualitative Analysis}
We visualize the class activation heatmaps in Figure~\ref{sf1} from the output foreground class token to the RGB image by using Grad-CAM~\cite{cam} visualization. We primarily show the qualitative visualization results for the LAVISH~\cite{lavish}, MBT~\cite{mbt}, and proposed MA-AVT models. We note that the proposed modality alignment techniques better discover target semantic regions than competitive baselines in general. Also, MA-AVT generates sharper boundaries with more weights on the target semantic regions representing sounding objects. Moreover, MA-AVT significantly reduces the attention weights in silent regions of images compared to other baselines. We hypothesize that such effective localization of the audio-visual semantic regions leads to its superior performance on audio-visual recognition tasks. 

\section{Conclusion}
In this paper, we present modality alignment techniques for parameter-efficient audio-visual transformer, dubbed MA-AVT. Our approach learns unimodal and multimodal tokens to adapt to unique unimodal features and  extract common multimodal features thereby achieving superior performance. The local self-attention modules are found to be effective with learnable tokens to focus on the most relevant tokens in each modality. To better contrast with the mismatched background scenarios, we introduce robust foreground mining that differentiates the corner case of complete modality mismatch.  We propose semantic contrastive learning to contrast across the semantic regions of each modality by utilizing the shared multimodal token embedding for achieving higher accuracy than baseline approaches. Moreover, we leverage blockwise contrastive learning for deeper alignment of cross-modal features for achieving consistent performance improvements. Modality-alignment training also demonstrates its effectiveness in unimodal test scenarios by considerably improving performance. Extensive experiments on three benchmark datasets show the superiority of the proposed MA-AVT over state-of-the-art methods.

\subsection*{Acknowledgements}
This research was supported in part by ONR Minerva program, iMAGiNE - the Intelligent Machine Engineering Consortium at UT Austin, and a UT Cockrell School of Engineering Doctoral Fellowship.

{
    \small
    \bibliographystyle{ieeenat_fullname}
    \bibliography{ref}
}

%\appendix
%\input{supp}

\end{document}